\documentclass[10pt,twocolumn,letterpaper]{article}

 \usepackage{cvpr}              

\definecolor{cvprblue}{rgb}{0.21,0.49,0.74}
\usepackage[pagebackref,breaklinks,colorlinks,allcolors=cvprblue]{hyperref}

 \usepackage{booktabs}
 \usepackage{graphicx}
 \usepackage{multirow}
 \usepackage[export]{adjustbox}
\usepackage{soul}
\usepackage{amssymb}

\usepackage{latexsym}
\usepackage{amsthm}
\usepackage{enumitem}
\usepackage{color}

\usepackage{graphicx}
\usepackage{amsmath}
\usepackage{amssymb}
\usepackage{booktabs}
\usepackage{algorithm}
\usepackage{algpseudocode}
\usepackage{multirow}

\newcommand{\mysection}[1]{\vspace{2pt}\noindent\textbf{#1}}
\usepackage{amssymb}
\usepackage{pifont}
\newcommand{\cmark}{\ding{51}}%
\newcommand{\xmark}{\ding{55}}%

\usepackage[dvipsnames]{xcolor}
\usepackage{caption}
\definecolor{delectricblue}{RGB}{93, 117, 131}
\colorlet{lightdelectricblue}{delectricblue!50} 
\definecolor{cambridgeblue}{rgb}{0.64, 0.76, 0.68}
\definecolor{bluegray}{rgb}{0.4, 0.6, 0.8}
\colorlet{delectblue}{bluegray!50}
\colorlet{delectgreen}{cambridgeblue!50}

\definecolor{mintgreen}{RGB}{193, 225, 193}
\definecolor{lightblue}{RGB}{173, 216, 230}
\definecolor{lightgreen}{RGB}{144, 238, 144}
\definecolor{lightyellow}{RGB}{255, 255, 204}
\definecolor{lightorange}{RGB}{255, 200, 150}
\definecolor{lightpurple}{RGB}{216, 191, 216}
\definecolor{lightcyan}{RGB}{224, 255, 255}
\definecolor{lightgray}{RGB}{230, 230, 230}

\usepackage{setspace}

\definecolor{cvprblue}{rgb}{0.21,0.49,0.74}
\usepackage[pagebackref,breaklinks,colorlinks,allcolors=cvprblue]{hyperref}

\def\paperID{19} 
\def\confName{CVPR}
\def\confYear{2025}

\begin{document}
\title{SSLR: A Semi-Supervised Learning Method for Isolated Sign Language Recognition}



\author{
\makebox[\textwidth][c]{%
    \parbox{17cm}{\centering
    Hasan Algafri$^{1}$, Hamzah Luqman$^{1,2}$, Sarah Alyami$^{1,3}$, and Issam Laradji$^{4}$  \\
    $^1$Information and Computer Science Department, King Fahd University of Petroleum and Minerals\\
    $^2$SDAIA–KFUPM Joint Research Center for Artificial Intelligence\\
    $^3$Computing Department, Imam Abdulrahman Bin Faisal University\\
    $^4$ServiceNow \\
    {\tt\small\{s201817820, hluqman\}@kfupm.edu.sa,  snalyami@iau.edu.sa,  issam.laradji@servicenow.com}
    }
}
}

\maketitle



\begin{abstract}
Sign language is the primary communication language for people with disabling hearing loss. Sign language recognition (SLR) systems aim to recognize sign gestures and translate them into spoken language. One of the main challenges in SLR is the scarcity of annotated datasets. To address this issue, we propose a semi-supervised learning (SSL) approach for SLR (SSLR), employing a pseudo-label method to annotate unlabeled samples. The sign gestures are represented using pose information that encodes the signer's skeletal joint points. This information is used as input for the Transformer backbone model utilized in the proposed approach. To demonstrate the learning capabilities of SSL across various labeled data sizes, several experiments were conducted using different percentages of labeled data with varying numbers of classes. The performance of the SSL approach was compared with a fully supervised learning-based model on the WLASL-100 dataset. The obtained results of the SSL model outperformed the supervised learning-based model with less labeled data in many cases.  

\end{abstract}    

\section{Introduction}
\label{sec:intro}

Sign language is the primary communication medium for people with disabling hearing loss. According to the World Health Organization\footnote{https://www.who.int/health-topics/hearing-loss}, hearing loss affects over 5\% of the world's population, and this number is estimated to be around 10\% by 2050. The growing population of hearing impaired people emphasizes the urgency of developing systems to support their integration into society.

Sign language is a descriptive language that depends on hands and body motions to describe objects and convey meaning \cite{suliman2021arabic}. It also combines facial expressions with manual gestures to represent some linguistics that could not be represented using manual gestures \cite{podder2023signer}. Sign language is an independent natural language with grammar and structure that differ from spoken languages \cite{alyami2023isolated}. There are several sign languages, such as American, British, Chinese, and Arabic. Each of these languages has a vocabulary that differs from other sign languages. 


Sign language recognition (SLR) aims to recognize sign gestures and translate them into spoken language. SLR systems can be classified based on the recognized signs into isolated and continuous \cite{el2022comprehensive}. Isolated SLR systems assume one sign is available in the sign video, whereas continuous SLR systems recognize continuous sign language sentences. 
  Moreover, SLR approaches fall into two categories vision-based and pose-based \cite{rastgoo2021sign}. Vision-based approaches recognize sign gestures 
  displayed on images or video streams \cite{9681230, Zhang2021}, while pose-based approaches employ the hand and body pose information as the main source of information for recognizing signs \cite{Bohacek2022, Tunga2021, alyami2023isolated, Bencherif2021}. Although vision-based models have achieved promising results, the trained models usually overfit the signer which can justify the degradation in the recognition rate when evaluating these systems on unseen signers. In contrast, pose-based systems deal with the signer pose information, therefore, changing the environment or signer has a small effect on the trained models. 
  In addition, processing poses information requires less computational power compared with video streams \cite{Bohacek_2022_WACV}. 
  To leverage the benefits of vision and pose techniques and to obtain more comprehensive representations, several researchers leveraged both sign video and pose information for SLR \cite{chen2023twostream,Aditya2022,Zhou2022}.

One of the main challenges for SLR is the lack of an annotated dataset \cite{el2022comprehensive}. Despite the availability of numerous sign language videos on social media or YouTube, these samples are mostly unannotated, limiting their use with supervised learning-based techniques. 
This challenge has motivated researchers to explore alternative learning approaches that can leverage unlabeled data for training SLR models. One of these approaches is semi-supervised learning (SSL), which utilizes both labeled and unlabeled data to train SLR models. The training process typically involves using labeled data to guide the model's learning process in labeling unannotated samples \cite{singh2021semi}. Samples labeled with high confidence are subsequently added to the labeled dataset and used to retrain the model.

In this study, we present a pose-based semi-supervised sign language recognition (SSLR) system that employs a pseudo labeling-based approach for leveraging the unlabeled data. The proposed methodology utilizes a Transformer model as a backbone for sign recognition and classification. Various proportions of labeled data were employed to train the transformer-based model. The trained model is subsequently utilized to classify unlabeled data. Sign samples recognized with high confidence are selected for inclusion in the labeled dataset.
This expanded labeled dataset is then used to re-train the model, and this iterative process is repeated until all samples are annotated. The proposed approach is evaluated using the WLASL dataset \cite{Li2020}, which consists of 100 classes. Promising results were obtained from this approach, particularly when compared to a fully supervised learning (FSL) model employing the same ratio of labeled data. 

Our contributions can be summarized as follows:

\begin{itemize}
    \item We proposed a semi-supervised learning method called SSLR designed for sign language recognition.
    \item We efficiently encoded pose information from sign gestures using the signer’s skeletal joint points, along with effective normalization and augmentation techniques.
    \item  We showed that SSLR significantly outperforms fully supervised methods, even with the same amount of labeled data.
\end{itemize}

\section{Related Work}

Several techniques have been proposed in the literature for SLR. These techniques can be categorized into supervised learning and SSL-based techniques.

\mysection{Supervised learning for SLR.} 
The emergence of powerful deep learning architectures has encouraged researchers to utilize convolutional neural networks (CNNs) followed by recurrent neural networks (RNNs)  for SLR. CNNs are employed to extract spatial representations from video frames \cite{Luqman2021,Rastgoo2020, Zhang2021}, while RNNs are utilized for temporal learning \cite{Luqman2021, Rastgoo2020a, Aly2020,Suliman2021,Lee2021, Chaikaew2021}. Other studies utilized the capabilities of 3D-CNN to simultaneously capture spatio-temporal information \cite{Huang2019, Rastgoo2020, jiang2021skeleton, Zhang2021}.  The Inflated 3D ConvNet (I3D) model \cite{carreira2017quo}, originally proposed for action recognition, was leveraged in several SLR frameworks  \cite{Li2020, Joze2020, Adaloglou2021}. The comparative analysis \cite{Li2020} demonstrated the superiority of I3D over a Graph Convolution Network (GCN) based model for SLR. 
Ozdemir et al. \cite{ozdemir2023multi} presented a pose-based GCN model, which integrated the GCN with multi-cue long short-term memory networks (MC-LSTMs) to model cues from hands, body, and face. The model obtained 90.85\% accuracy on the AUTSL dataset with 100 classes.


The success of Transformer networks in various domains has encouraged researchers to develop Transformer-based SLR frameworks \cite{Bohacek2022,Camgoz2020,DeCoster2020,Aloysius2021,Selvaraj2021,Zhou2021a,Tunga2021}. De Coster et al. \cite{DeCoster2020} conducted a comparative analysis of RNNs, Transformers, Pose-based Transformers, and multi-modal Transformer networks.  The multi-modal framework was fed with the pose information and features extracted using 2DCNN. It obtained the best performance with 74.70\% accuracy. Tunga et al. \cite{Tunga2021} combined a GCN and a BERT-based model to encode signer pose data, obtaining 60.1\% accuracy on the Word-Level American Sign Language (WLASL) dataset. 

Although high recognition accuracies were reported using FSL models, the need for large annotated datasets limits the usage of these techniques for many sign languages. According to \cite{el2022comprehensive}, most of the SLR studies targeted American \cite{li2020word} and Chinese sign languages \cite{rastgoo2021sign}. This can be attributed to the availability of large annotated datasets for these languages. However, this is not the case for most sign languages where resources are scarce. Moreover, the collection and annotation of SLR datasets is time-consuming and expensive due to the need of employing sign language experts. 

\mysection{Semi-supervised learning for SLR.}
Low-data learning approaches can potentially address the challenge of limited data availability for some sign languages. These approaches involve training models with a small amount of labeled data using techniques, such as transfer learning and data augmentation, to enhance model performance. WiSign \cite{shang2017robust} adopted a co-training SSL approach using  support vector machine (SVM) and KNN,  The classifiers were trained on labeled data and were utilized to predict possible labels for the unlabeled instances. If both classifiers predict the same label for a particular instance, it is labeled as such and added to the labeled dataset. This iterative process continues until all instances are labeled. 

Selvaraj et al. \cite{selvaraj2021openhands} explored transfer learning from Indian sign language to other languages with self-supervised pretraining strategies. The results revealed that cross-lingual transfer learning is effective and led to accuracy gains ranging from 2\% up to 18\% in low-resource languages. The authors also presented OpenHands, an SLR library that offers a variety of pre-trained models based on pose data. Four architectures were included in this library: LSTM, Transformer, and two graph-based models. Cross-lingual transfer learning was also explored in \cite{bird2020british}. The study presented a late fusion model for British SLR. RGB input is modeled with VGG16-MLP, and Leap data are modeled with evolutionarily optimized deep MLP. Late fusion is employed to predict the final sign class. The study demonstrated the efficacy of transfer learning from British to American sign languages. The multi-modal approach was evaluated on a small dataset consisting of 18 signs collected for this study and achieved an accuracy of 94.44\%.
An interesting recent study \cite{bilge2022towards} investigated zero-shot learning for SLR, where a model is trained to detect signs not included in the training dataset. The framework consisted of four data streams. Two streams for visual description were extracted from sign videos, and two textual streams include a detailed text description of the signs obtained from the Webster sign language and the American sign language handshape dictionaries. The study explored various models for spatio-temporal video modeling and employed a BERT architecture for textual modeling. While the experimental results demonstrated promising potential for zero-shot recognition of signs, the achieved accuracy levels were relatively modest compared to other domains leveraging zero-shot learning.

\section{Methodology}

\begin{figure*}[t]
  \centering
   \includegraphics[width=0.9\linewidth]{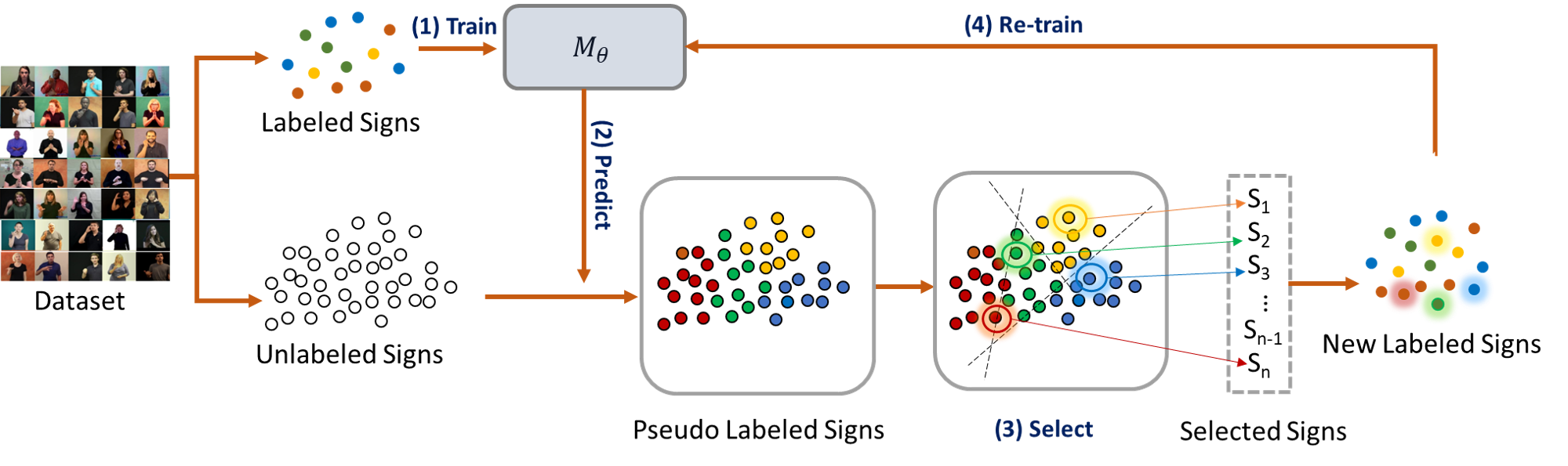}
   \caption{\textbf{The pipeline of the proposed SSL model.} The model is (1) trained on labeled signs using the pose-based transformer model. This model is used later to (2) predict signs of the unlabeled samples. Then, signs of unlabeled samples predicted with high confidence are (3) selected and added to the labeled set. This updated set is used to (4) re-train the model. This process is repeated until all unlabeled samples are labeled.}
   \label{fig:framework}
\end{figure*}

 Figure \ref{fig:framework} illustrates the high-level flow of the proposed SSL framework. The proposed model processes signs represented through their pose information, which encodes the signer's hands, body, and face using a set of joint points.  This information helps in tackling image-related challenges by eliminating the need for image pre-processing and the detection and localization of signer hands and body. In addition,  the pose information plays an important role in the development of signer-independent recognition systems, as this information encodes the signer's skeleton and does not overfit the identity of the signer. To train the SSL model, the dataset signs were split into annotated and unannotated signs with different percentages. The annotated set was used to train a Transformer model, which would later be employed to classify the unannotated samples. A pseudo-labeling SSL method was employed to label the unlabeled samples. Subsequently, samples with high confidence were extracted from the unannotated set and added to the labeled signs for re-training the model. This process was iteratively repeated until all unannotated signs were labeled.     

\subsection{Preprocessing}\label{sec:preprocessing}

Several augmentation techniques have been employed to prevent model overfitting and enhance its generalization. These techniques are applied to the pose data in each sign frame. Initially, random Gaussian noise was added to the training data. Subsequently, in-plane rotation was applied to rotate joint points in each frame by an angle of up to 13 degrees, randomly selected \cite{Bohacek2022}. Another rotation was performed at the arm level, where the joint points of each arm were slightly rotated by an angle of $\pm$4 degrees with respect to the other arm. Additionally, data shearing was performed by horizontally and vertically squeezing each frame from both sides. Each frame was randomly squeezed by 15\% from the left and right sides.

The coordinates of each landmark in the pose information depend on the recording settings, such as the distance from the camera and tilting. These values can results in more training time and can result in learning spatial features unrelated to the recorded signs \cite{Bohacek2022}. To mitigate this issue, we adopted the normalization technique proposed in \cite{bauer2014use}. This technique normalizes the pose information by projecting it into the signing space. The signer's body parts are normalized relative to the signer's head height, while the coordinates of the hands are normalized based on their bounding boxes.

\subsection{Backbone Model}
\begin{figure}[t]
  \centering
   \includegraphics[width=0.6\linewidth]{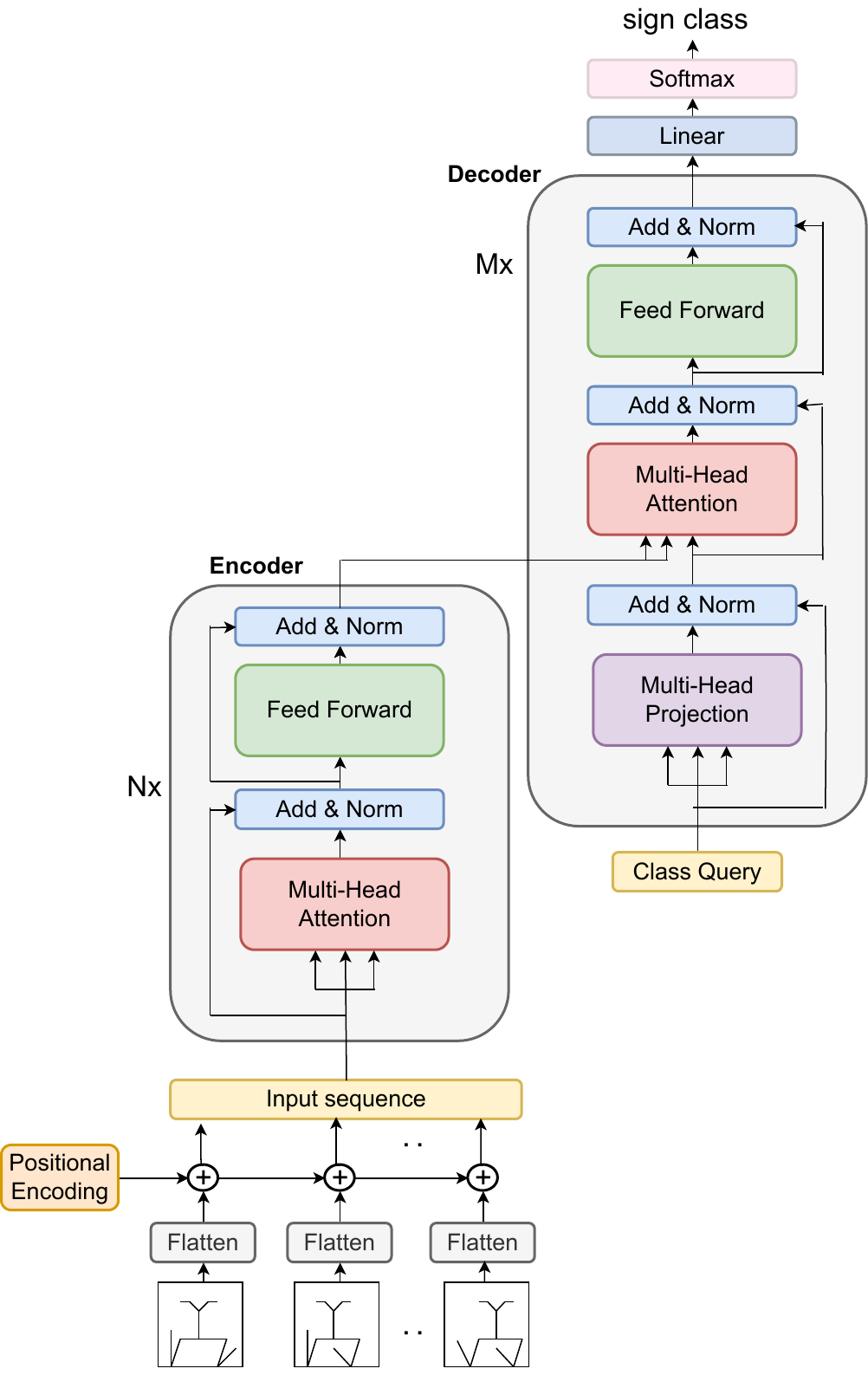}
   \caption{Architecture of the utilized Transformer model \cite{Bohacek2022}. }
   \label{fig:spoter}
\end{figure}

We used the architecture of a proposed system for word-level sign language recognition called SPOTER \cite{Bohacek2022}. The model is based on a Transformer with slight modifications, as shown in Figure \ref{fig:spoter}. It takes the pose coordinates of 54 body joints as input, comprising a 108 sized input vector for each image. Positional information is first embedded into the input vector. The encoder consists of 6 blocks, each equipped with a self-attention module having 9 heads and a feed-forward network. 

The Transformer has 6 decoder blocks. The decoder receives a single query for the sign class at the input, referred to as the "Class Query." A multi-head projection module is employed, designed for sequences with only one element. Therefore, only the projection of the input vector into the value space is utilized. Multiple parallel projection heads are retained, and their projections are combined and processed by the final linear layer within the hidden dimension. Subsequently, the output from the encoder is combined with the projected class query in the decoder multi-head attention module. The decoded class query is fed into a classification layer with softmax activation to produce the sign label.

\subsection{Semi-supervised learning Sign Language Recognition (SSLR)}
This work uses pseudo-labeling SSL method for labeling unannotated samples. 
Pseudo-labeling is an SSL approach that involves using a model trained on labeled data to generate labels for unlabeled data. Figure \ref{fig:framework} and Algorithm \ref{alg:algo} illustrate the framework of the proposed model. 
The proposed approach involves four main steps: initial training, predicting unlabeled data, incorporating pseudo-labels, and re-training.

The used dataset consists of two disjoint subsets, a small labeled data $L = ((x_i,y_i); i \in (1, ... ,N$)), and a larger unlabeled set $U = (x_j; j \in (1, ... ,K$)). where $x$ is the input sample and $y$ is its label which is the equivalent sign of that sample. 
The labeled set $L$ is used in the \textit{initial training} to train the Transformer model $M$. This model is trained with the Cross-Entropy loss function and Stochastic Gradient Descent optimizer. The trained transformer model is then used in the following stage to predict and infer pseudo-labels for the unlabeled samples $U$. Each sample is assigned a set of probabilities representing the likelihood of classifying the sample into the corresponding sign's class. For each class, the pseudo-labeled sample with the highest confidence is selected. The unlabeled data is augmented with the pseudo-labels generated by the model, creating a larger dataset that now includes both the original labeled data and the newly pseudo-labeled data. The model is subsequently re-trained on the combined dataset. This process is repeated until all samples in the unlabeled set have been labeled. The training algorithm is shown in Algorithm \ref{alg:algo}.

\begin{algorithm}
\caption{SSL with Pseudo Labeling}\label{alg:algo}
\begin{algorithmic}[1]
\State \textbf{Dataset:} Labeled data $L = ((x_i,y_i); i \in (1, ... ,N$)), Unlabeled data $U = (x_j; j \in (1, ... ,K$))
\State $M.fit(L)$ \Comment{Initialize model $M$ by training on $L$.}
\Repeat
\State $P = M.predict(U)$ \Comment{Generate pseudo labels $P$}
\State $\hat{P} = max(P)$ \Comment{Get $P$ with the highest confidence}
\State $U = U - \hat{P}$ \Comment{Remove $\hat{P}$ from $U$}
\State $L = L \cup \hat{P}$ \Comment{Add $\hat{P}$ to $L$}
\State $M.fit(L)$ \Comment{Re-train on the new labeled data}
\Until{convergence or maximum iterations are reached.}
\end{algorithmic}
\end{algorithm}

\section{Experimental Work}
\mysection{Dataset.}
We used the American Sign Language (WLASL) video dataset \cite{Li2020} to train and evaluate the proposed approach. The dataset consists of 100 signs performed by several signers. The signs were collected in a constrained environment where a unified background was used. To address the variability between signers, we focused on the pose information that encodes the signer's skeleton joint points coordinates in each sign frame. The signer's face is also represented using a set of landmarks that help capture the facial expressions used with sign gestures. The dataset is split into train, validation, and test sets following the same ratio of 4:1:1 used by the dataset authors.

\mysection{Results and Discussion.}
Several experiments have been conducted to evaluate the proposed approach. The experiments were conducted by training two models: an FSL model that was trained using the labeled data and an SSL that utilized both labeled and unlabeled data. The FSL utilizes the Transformer model which is also used as a backbone of the SSL model. We used the results of the FSL as a baseline to compare the performance of the proposed SSL approach.

 Table \ref{tab:results} shows the results obtained from the FSL and SSL approaches on 100 classes of the WLASL dataset. Throughout these experiments, we varied the portion sizes of labeled data used to train the SSL model, enabling an evaluation of the effect of labeled data size on overall model performance. Results for the FSL model, using the same labeled data portions, are also presented. As shown in the table, our proposed SSL model either closely approximates or outperforms the FSL approach across the majority of labeled data proportions. Furthermore, increasing the size of labeled data from 25\% to 50\% enhanced the SSL model's accuracy by 15.5\%, in contrast to an 11.6\% improvement observed in the FFL model. This highlights the SSL model's effectiveness in utilizing the expanded labeled dataset to annotate unlabeled data, thereby significantly boosting recognition accuracy. Furthermore, training the model with a small portion of labeled data requires less training time, giving the SSL approach an advantage over the FSL approach.

\begin{figure}[t]
  \centering
   \includegraphics[width=\linewidth]{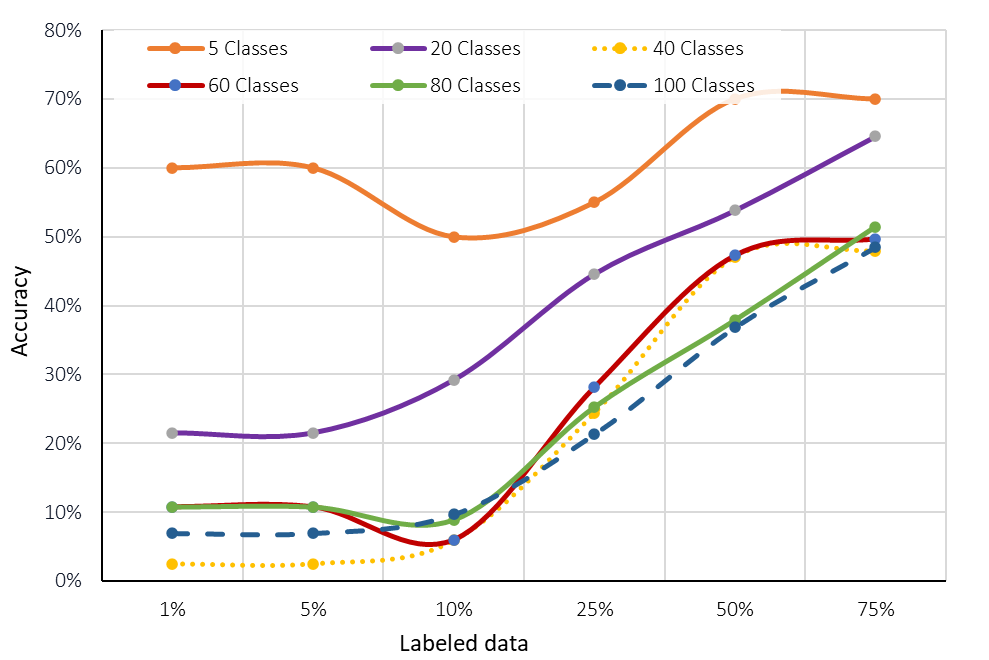}
   \caption{The accuracies of the SSL model with different number of classes. The x-axis represents the percentage of labeled data used to train the model.}
   \label{fig:classes_results}
\end{figure}

To evaluate the generalization of the proposed model, we present results for different numbers of sign classes in  Table \ref{tab:classes} and Table \ref{fig:classes_results}. As shown in the table, the SSL model demonstrates robust generalization across different class sizes. Although there is a reduction in the accuracy when increasing the number of classes, this reduction can be attributed mainly to the increase in the number of evaluated samples and the expanded sign space. This behavior is also followed by the FSL model. 
For a small number of classes (5 classes), the FSL model's performance was notably low when trained with limited labeled data, while the SSL model was able to learn even with just 1\% of the labeled data. As both models were evaluated on a larger number of classes, the SSL model consistently outperformed the FSL, especially when trained with 75\% of the labeled data.


\begin{table}[h]
  \centering
   \caption{The accuracy of FSL and SSL on 100 classes with different percentages of labeled data.}
  \label{tab:results}
 \begin{tabular}{lcccccc}
    \toprule
    
    Labeled Data & 1\% & 5\% & 10\% & 25\% & 50\% & 75\% \\ \midrule
    
    FSL &  7.0 &    7.0 &   8.1&   22.5 & 35.3 &  48.1   \\ \midrule
    
    SSL (ours) &  7.0 &    7.0 &   9.7 & 21.3 &  36.8 &  48.4  \\
    \bottomrule
  \end{tabular}

\end{table}
\begin{table*}[]
\centering
\caption{The accuracy of the FSL and SSL with a different number of classes. }
\label{tab:classes}
\begin{tabular}{c|cc|cc|cc|cc|cc|cc}
\hline
\multirow{3}{*}{\textbf{Labeled   data}} & \multicolumn{12}{c}{\textbf{Number   of classes}}                                                                                                                                                                                                                                          \\ \cline{2-13} 
                                         & \multicolumn{2}{c|}{5}                           & \multicolumn{2}{c|}{20}                          & \multicolumn{2}{c|}{40}                          & \multicolumn{2}{c|}{60}                          & \multicolumn{2}{c|}{80}                          & \multicolumn{2}{c}{100}
                                         \\ \cmidrule{2-13} 
                                         
                                         & \textbf{FSL} & \multicolumn{1}{c|}{\textbf{SSL}} & \textbf{FSL} & \multicolumn{1}{c|}{\textbf{SSL}} & \textbf{FSL} & \multicolumn{1}{c|}{\textbf{SSL}} & \textbf{FSL} & \multicolumn{1}{c|}{\textbf{SSL}} & \textbf{FSL} & \multicolumn{1}{c|}{\textbf{SSL}} & \textbf{FSL} & \textbf{SSL} \\ \hline
\textbf{1\%}                             & 25.0         & \multicolumn{1}{c|}{60.0}         & 26.2         & \multicolumn{1}{c|}{21.5}          & 2.5          & \multicolumn{1}{c|}{2.5}          & 12.0         & \multicolumn{1}{c|}{10.8}         & 9.8         & \multicolumn{1}{c|}{10.7}          & 7.0          & 7.0          \\
\textbf{5\%}                             & 25.0         & \multicolumn{1}{c|}{60.0}         & 29.2         & \multicolumn{1}{c|}{21.5}         & 2.5          & \multicolumn{1}{c|}{2.5}          & 12.0         & \multicolumn{1}{c|}{10.8}         & 9.8         & \multicolumn{1}{c|}{10.7}          & 7.0          & 7.0          \\
\textbf{10\%}                            & 65.0         & \multicolumn{1}{c|}{50.0}         & 24.6         & \multicolumn{1}{c|}{29.2}         & 6.7          & \multicolumn{1}{c|}{5.9}          & 10.8         & \multicolumn{1}{c|}{6.0}          & 8.9          & \multicolumn{1}{c|}{8.9}          & 8.1          & 9.7          \\
\textbf{25\%}                            & 60.0         & \multicolumn{1}{c|}{55.0}         & 46.2         & \multicolumn{1}{c|}{44.6}         & 22.7         & \multicolumn{1}{c|}{24.4}         & 32.3         & \multicolumn{1}{c|}{28.1}         & 29.0         & \multicolumn{1}{c|}{25.2}         & 22.5         & 21.3         \\
\textbf{50\%}                            & 60.0         & \multicolumn{1}{c|}{70.0}         & 60.0         & \multicolumn{1}{c|}{53.8}         & 36.1         & \multicolumn{1}{c|}{47.1}         & 46.7         & \multicolumn{1}{c|}{47.3}         & 45.8         & \multicolumn{1}{c|}{37.9}         & 35.3         & 36.8         \\
\textbf{75\%}                            & 65.0         & \multicolumn{1}{c|}{70.0}         & 61.5         & \multicolumn{1}{c|}{63.1}         & 42.9         & \multicolumn{1}{c|}{47.9}         & 50.9         & \multicolumn{1}{c|}{49.7}         & 45.3         & \multicolumn{1}{c|}{51.4}         & 48.1         & 48.4         \\ \hline
\end{tabular}

\end{table*}

\begin{figure}[t]
  \centering
   \includegraphics[width=\linewidth]{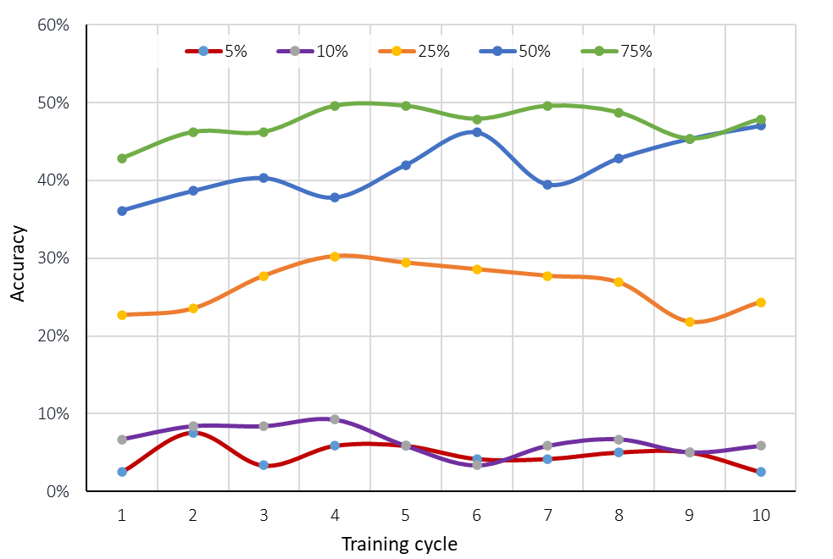}
   \caption{The performance of the SSL model on 40 classes with different percentages of labeled data. The horizontal axis represents the training cycle that involves an incremental increase of the pseudo-labeled samples.}
   \label{fig:cycles}
\end{figure}

One of the stages in SSL training involves iterative pseudo-labeling.  Figure \ref{fig:cycles} illustrates the impact of incrementally incorporating more pseudo-labeled samples into the labeled set during the SSL model's training on 40 classes. In each cycle, a set of unlabeled samples is labeled and added to the labeled data for subsequent model re-training. As depicted in the figure, the performance of the SSL model converges early when trained on smaller percentages of the labeled data, and the addition of more pseudo-labeled samples does not lead to further improvements. In contrast, the performance of the SSL model continues to improve when trained on larger labeled data sizes as more pseudo-labeled samples are added. This improvement can be attributed to the accuracy of labeling unlabeled samples through the pseudo-labeling method. Training the model on larger labeled data sizes results in a more accurate classification of unlabeled data, thereby positively improving the overall accuracy of the model.

\subsection{Ablation study}

An ablation study was conducted to evaluate the importance of the preprocessing and augmentation techniques used with the proposed SSL approach. We evaluated these techniques on 20 classes with 75\% labeled data and the obtained results are shown in Table \ref{tab:augmentation}.

\mysection{Normalization. } 
The pose information was normalized by projecting it into the signing space, as discussed in Section \ref{sec:preprocessing}. To assess the effectiveness of normalization for SLR, we evaluated the proposed SSL model with and without normalization. As indicated in Table \ref{tab:augmentation}, training the model in the raw pose data without normalization resulted in a test accuracy of 46.2\%, whereas normalizing these data increased the recognition accuracy by more than 10\%. When the model is fed with raw pose data, it tends to learn spatial features closely tied to the recording environment, such as the distance from the camera and the signer's position within the frame.  In contrast, normalization alleviated these issues by making the coordinates of the joint points relative to the signer's body.
 
\mysection{Augmentation. }
Several augmentation techniques have been applied to enrich the labeled set and prevent the model's overfitting, as discussed in Section \ref{sec:preprocessing}.  Augmentation played a role in enhancing the test accuracy by around 5\%, as demonstrated in Table \ref{tab:augmentation}. Both rotation and shearing had a significant impact on improving the accuracy of the SSL model.

\begin{table}[]
\caption{Ablation study results with normalization and different augmentation techniques.}
\label{tab:augmentation}
\resizebox{\linewidth}{!}{\begin{tabular}{cccccc}
 \toprule
\textbf{Shear} & \textbf{Rotation} & \textbf{Gaussian Noise} & \textbf{Normalization} & \textbf{Val Acc.} & \textbf{Test Acc.} \\ \midrule
\textcolor{black}{ \xmark }        &      \textcolor{black}{ \xmark }      &            \textcolor{black}{ \xmark }         &               \textcolor{black}{ \xmark }                       & 29.8                  & 46.2                   \\  
\textcolor{black}{ \xmark }       &       \textcolor{black}{ \xmark }      &            \textcolor{black}{ \xmark }         &               \textcolor{black}{ \cmark }                       & 61.9                  & 58.5                   \\  
\textcolor{black}{ \xmark }   &           \textcolor{black}{ \xmark }      &            \textcolor{black}{ \cmark }       &                 \textcolor{black}{ \cmark }                       & 58.3                  & 56.9                   \\  
\textcolor{black}{ \xmark }    &          \textcolor{black}{ \cmark }    &              \textcolor{black}{ \cmark }     &                   \textcolor{black}{ \cmark }                       & 60.7                  & 58.5                   \\  
\textcolor{black}{ \cmark }         &       \textcolor{black}{ \cmark }        &          \textcolor{black}{ \cmark }           &             \textcolor{black}{ \cmark }                       & 63.1                  & 63.1                   \\ \bottomrule
\end{tabular}}

\end{table}

\section{Conclusion}
Sign language serves as the primary communication language for deaf and hard-of-hearing individuals. To bridge the communication gap with the hearing community, various models have been developed in the literature. However, the lack of annotated data remains a significant challenge in building reliable SLR systems. This paper aims to address the shortage of resources for SLR by proposing an SSL approach using the pseudo-labeling method. We utilized pose data to train a Transformer model that was used as the backbone model for the proposed approach. The evaluation of our approach was conducted on the WLASL dataset, which consists of 100 classes. We compared the performance of the proposed SSL approach with a supervised learning-based model. To showcase the learning capabilities of SSL across different data sizes, we conducted our experiments using various percentages of labeled data. Additionally, we varied the number of classes to assess the model's generalization, and the reported results either closely matched or outperformed the supervised learning-based model in many cases. These results demonstrate the potential of this approach for developing SLR systems. Future work can investigate improving the labeling process through pseudo-label refinement techniques such as uncertainty-thresholding. Also, adaptive class-balanced pseudo-labeling can be investigated to overcome unbalanced class pseudo-labels.

\section*{Acknowledgment}
 The authors would like to acknowledge the support received from the Saudi Data and AI Authority (SDAIA) and King Fahd University of Petroleum and Minerals (KFUPM) under the SDAIA-KFUPM Joint Research Center for Artificial Intelligence Grant no. JRC-AI-RFP-14.



{
    \small
    \bibliographystyle{ieeenat_fullname}
    \bibliography{main}
}


\end{document}